\documentclass[a4paper, 10pt, conference]{ieeeconf}      
\usepackage{tabu}                      
\usepackage{booktabs}                  
\usepackage{lipsum}                    
\usepackage{mwe} 
\usepackage{FG2025}
\usepackage[ruled,vlined]{algorithm2e}
\usepackage{amsmath}
\usepackage{tikz}
\usepackage{multirow}
\usetikzlibrary{positioning}
\usetikzlibrary{calc}
\usepackage{mathptmx}                  
\usepackage{float}
\usepackage{algorithm2e}
\usepackage{tikz}
\usepackage{graphicx}

\usepackage{subcaption}
\usepackage{marvosym} 

\usetikzlibrary{positioning, calc, shapes, arrows.meta}
\FGfinalcopy 

\IEEEoverridecommandlockouts                              
\def\FGPaperID{220} 

\title{\LARGE \bf
Duo Streamers: A Streaming Gesture Recognition Framework
}

\author{\parbox{16cm}{\centering
    {Boxuan Zhu$^{1,2}$ \texttt{<boxuan.zhu@liverpool.ac.uk>}, \\Sicheng Yang$^1$ \texttt{<sicheng.yang@xintelligencelabs.ac>}, \\Zhuo Wang$^1$ \texttt{<zhuo.wang@xintelligencelabs.ac>}, \\Haining Liang$^{3}$ \texttt{<hainingliang@hkust-gz.edu.cn>}, \\Junxiao Shen$^{4, \textrm{\Letter}}$\thanks{\footnotesize Corresponding author} \texttt{<junxiao.shen@bristol.ac.uk>}}\\
    \vspace{0.5cm} 
    {\normalsize
    $^1$ X-Intelligence Labs 
    $^2$ University of Liverpool 
    $^3$ HKUST (Guangzhou) 
    $^4$ University of Bristol \\
    }
}}

\begin{document}

\twocolumn[{%
\renewcommand\twocolumn[1][]{#1}%
\graphicspath{{figs/}{figures/}{pictures/}{images/}{./}}
\ifFGfinal
\thispagestyle{empty}
\pagestyle{empty}
\else
\author{Anonymous FG2025 submission\\ Paper ID \FGPaperID \\}
\pagestyle{plain}
\fi
\maketitle

\begin{center}
    \centering
    \captionsetup{type=figure}
    \includegraphics[width=.8\textwidth,height=5.5cm]{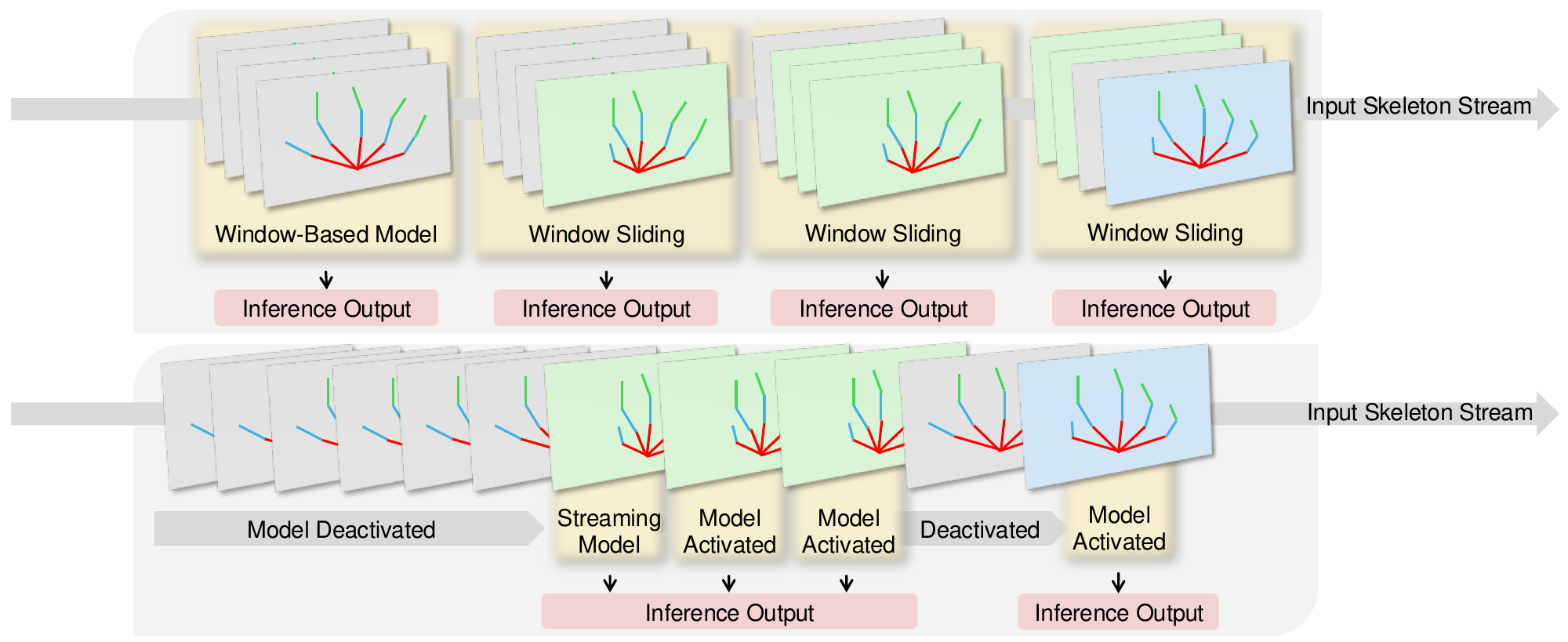}
    \captionof{figure}{Window-based Model vs. Streaming Model. The proposed streaming model reduces model size and can be activated on demand to deliver immediate inference over the data stream, thereby lowering latency and improving real‐time performance.}
\end{center}%
}]


\begin{abstract}

Gesture recognition in resource-constrained scenarios faces significant challenges in achieving high accuracy and low latency. The streaming gesture recognition framework, Duo Streamers, proposed in this paper, addresses these challenges through a three-stage sparse recognition mechanism, an RNN-lite model with an external hidden state, and specialized training and post-processing pipelines, thereby making innovative progress in real-time performance and lightweight design. Experimental results show that Duo Streamers matches mainstream methods in accuracy metrics, while reducing the real-time factor by approximately 92.3\%, i.e., delivering a nearly 13-fold speedup. In addition, the framework shrinks parameter counts to 1/38 (idle state) and 1/9 (busy state) compared to mainstream models. In summary, Duo Streamers not only offers an efficient and practical solution for streaming gesture recognition in resource-constrained devices but also lays a solid foundation for extended applications in multimodal and diverse scenarios. Upon acceptance, we will publicly release all models, code, and demos.

\end{abstract}


\section{INTRODUCTION}

Real-time gesture recognition in complex environments remains a significant challenge, especially on resource-constrained devices. With the widespread adoption of wearable devices, mobile terminals, and smart home systems, gesture recognition has become a key interface in human–computer interaction. However, systems often need to rapidly capture and recognize gestures under limited hardware resources and stringent power constraints, making it a pressing issue to balance accuracy, model size, and real-time inference. 

Window-based gesture recognition algorithms have made significant progress over the past decade; by continuously sliding a window across sequential frames, they perform feature extraction and classification, and thus exhibit stable performance on pre-segmented datasets \cite{luzhnica2016sliding}. However, when employed for real-time streaming inference, the frequent model invocation and the processing of numerous redundant frames lacking valid gestures can result in significant computational overhead \cite{10065484}. In response, researchers have begun exploring “streaming methods,” which feed data into the model in a continuous flow and require the model to generate predictions in real time as the data stream progresses \cite{yusuf2024real}, thereby eliminating the need for repeatedly buffering and processing large historical windows. While such approaches have garnered increasing attention in areas like network communications \cite{10464328} and financial transactions \cite{alam2024real}, they have not yet been widely adopted for the real-time analysis of physical actions such as gestures.

In the domain of gesture recognition, “streaming processing” means the model must immediately generate predictions as data arrive in real time, while avoiding the accumulation of latency caused by redundant computations. This requirement is even more critical for resource-constrained devices: if the model is too large or needs to buffer excessive raw inputs, the resulting memory overhead and energy consumption become prohibitive. Consequently, streaming methods give rise to three core demands: 
1) ultra-lightweight, 2) streaming processing, and 3) ultra-low latency
    
The first two demands focus on controlling model size and unnecessary computations, while the third emphasizes the importance of early recognition for real-time applications. With the continuing proliferation of wearable and mobile applications, these requirements apply equally to both high-performance devices and smaller-scale devices.

Against this backdrop, this paper introduces a new streaming gesture recognition framework called Duo Streamers. By adopting a simplified RNN-lite architecture and a three-phase recognition mechanism, Duo Streamers significantly reduces computational overhead without compromising accuracy, and supports continuous, immediate processing of streaming inputs. Compared to traditional approaches, this design is better suited for scenarios with sparse gesture occurrences: during idle periods, the system only needs to maintain minimal-power monitoring, and once a gesture is detected, it allocates its main computational resources to recognizing the key information—thus substantially reducing the overall burden. Additionally, Duo Streamers incorporates specialized training and post-processing pipelines, enabling direct training on pre-segmented datasets as a streaming model, while dynamically coordinating and adapting the framework during inference for further performance enhancements. As a result, Duo Streamers not only achieves extremely low latency on high-performance devices but can also be deployed on edge devices with limited computational capacity, thereby covering diverse interaction needs.

For experimental evaluation, we selected the SHREC2021 dataset \cite{caputo2021shrec} to validate the effectiveness of Duo Streamers. The results show that, under various conditions (whether gesture occurrences are frequent or extremely sparse), the framework maintains stable performance. Compared with baseline methods, Duo Streamers accelerates inference speed, reducing the real-time factor by approximately 92.3\% (nearly a 13-fold speedup), while simultaneously shrinking parameter counts to 1/38 in idle states and 1/9 in busy states.

In summary, to implement streaming methods and overcome real-time challenges in gesture recognition, our work presents three main innovations:
    \begin{enumerate}
        \item \textbf{Three-Stage Sparse Recognition Mechanism}: The framework comprises an ultra-lightweight binary detector, an ultra-lightweight multi-class recognizer, and a Euclidean analyzer. Through task division and interaction among these components, computational loads are distributed more evenly, significantly reducing redundancy during idle frames.
        \item \textbf{Stream-Based RNN-lite Model}: By storing critical temporal information in a highly compressed external hidden state, our method achieves both a lightweight design and robust sequential modeling. Both the Detector and the Recognizer within the framework utilize this architecture.
        \item \textbf{Innovative Model Training and Post-Processing Pipeline}: We devise both a streaming training pipeline and a post-processing pipeline tailored to the demands of training and deployment, ensuring that Duo Streamers adapts to diverse data sources and delivers additional performance gains after training.
    \end{enumerate}

\section{RELATED WORK}
In application scenarios such as virtual reality (VR) and augmented reality (AR), gesture recognition systems must capture and interpret gestures on edge devices as quickly as possible \cite{9283348, shen2024ringgesture}. This stringent real-time requirement arises not only from limited power and computing resources, but also from the need to simultaneously handle tasks such as spatial localization, motion tracking, and multi-channel interaction \cite{shen2023fast}. If the recognition process is delayed until after the gesture ends, users may experience discomfort, and the immersion and fluidity required for VR/AR interaction will be compromised \cite{shen2024towards}. Consequently, achieving stable real-time gesture recognition on these edge devices has become a pressing technical challenge. The following content briefly reviews algorithmic designs in the field of gesture recognition and extends to relevant real-time research.

\subsection{Deep Learning in Gesture Recognition}
Many technologies introduced from other domains into gesture recognition often overlook custom optimizations for edge devices and continue to rely on generalized model designs and neural network architectures. A common practice is to adopt window-based methods for processing continuous input \cite{8756576}, where raw frames are stacked over a certain time span to capture contextual information along the temporal dimension \cite{8756576, 9264164}. However, an excessively large window significantly increases model size and latency, while a window that is too small fails to provide sufficient temporal context; frame-based approaches, as an extreme case (window size of 1), are nearly incapable of recognizing dynamic gestures \cite{caputo2021shrec}. Moreover, for VR/AR devices requiring high-frequency interaction, pixel-based sliding window approaches impose additional costs for buffering raw video data, and this burden is projected to expand further with the advent of next-generation ultra-high-definition standards such as 8K.

At the algorithmic level, existing research primarily concentrates on CNN, LSTM, Transformer, and GNN architectures \cite{9873969}. While CNNs offer automatic feature extraction and often involve fewer parameters, convolutional operations can be time-consuming \cite{younesi2024comprehensive}. LSTMs excel at capturing temporal dependencies but typically rely on larger parameter sets \cite{justus2018predicting}. Transformers can model long-range dependencies and allow parallel computation, yet their computation remains considerable when handling linear, streaming inputs in real-world scenarios \cite{fournier2023practical}. GNNs are suitable for modeling skeletal data but demand manually designed preprocessing and graph generation pipelines, making direct adaptation to diverse settings challenging \cite{li2023graph}. Consequently, most of these methods still require multiple frames to be accumulated within a window for stable training and inference, falling short of the real-time requirements on edge devices.

Another challenge for real-time recognition lies in the need to repeatedly invoke the model on the input stream \cite{8578647, shen2024boosting}. When gestures are sparse and irregular, window-based recognition systems must frequently perform inference to align with potential incoming gestures \cite{caputo2021shrec}, which boosts recall but also raises false positive rates. To suppress false positives, some window-based approaches introduce lengthy “ignore periods” \cite{caputo2021shrec}, resulting in delayed recognition or even missed detections and conflicting with low-latency requirements. In scenarios demanding high recall, relying heavily on windows and frequent model invocation creates a dilemma between latency and accuracy.

In recent years, to balance performance and real-time needs, researchers have integrated deep learning with various data sources by leveraging ultrasound \cite{saez2021gesture, bimbraw2024forearm} or electromyography (EMG) \cite{kim2023emg} as event-driven inputs to support recognition, or adopting knowledge distillation, quantization, and pruning to deploy lightweight models \cite{bimbraw2024forearm}. Meanwhile, some studies focus on the robustness of actual deployments—considering factors such as environmental lighting, background complexity, and user variability \cite{murad2024advancements, shen2021imaginative}—and employ adaptive or transfer learning \cite{wu2024gesture, shen2024towards} to maintain accuracy and real-time performance. Overall, deep learning methods have already demonstrated strong capabilities in gesture recognition, including CNNs for visual feature extraction and LSTMs for capturing temporal dependencies \cite{li2021gesture, wu2024gesture}. Key directions for further improvement include multi-stream architectures \cite{rahim2024advanced, yaseen2024next, benitez2020finger}, lightweight designs \cite{bimbraw2024forearm, kim2023emg}, and adaptive technologies \cite{murad2024advancements, al2022structured}. Nonetheless, in high-interaction yet resource-constrained scenarios like VR/AR, it remains necessary to further reduce latency, power consumption, and storage requirements.

\subsection{Streaming Model}

Streaming methods prioritize real-time performance and have been extensively studied in areas such as speech recognition and autonomous driving, offering important references for gesture recognition. Unlike batch-based approaches, streaming models aim to produce predictions immediately upon data arrival, significantly reducing latency and enabling continuous online inference \cite{gomes2019machine, montiel2021river, lara2023data}.

In speech recognition, the Recurrent Neural Network Transducer (RNN-T) has been specifically optimized for low-latency scenarios by processing inputs incrementally, eliminating the need to wait for the complete sequence before generating outputs \cite{graves2012sequence}. Some variants of RNN-T also introduce boundary-aware training strategies that restrict the evaluation scope to key regions, reducing computational overhead and improving speed \cite{an2023bat}. Meanwhile, in streaming Attention Encoder-Decoder (AED) models, a technique known as monotonic attention sequentially focuses only on the relevant parts of the input, thus reducing repeated access to earlier segments and lowering computational load and latency \cite{chen2021developing}.

In the field of autonomous driving, some studies have adapted existing mature visual recognition schemes to implement streaming models. For instance, multispectral pedestrian detection based on the YOLOv4 architecture leverages lightweight design to achieve efficient computation and low latency \cite{s22031082}. In addition, for efficient motion prediction, researchers have introduced a simplified Transformer architecture—reducing pre-training complexity and combining simple linear and Transformer layers to shorten prediction time and cut resource consumption \cite{prutsch2024efficient}.

Similar approaches are gradually being applied to gesture recognition \cite{li2021gesture, devineau2018deep}, where models can parse incoming frames continuously upon data arrival, eliminating the need to wait for a full window and thus outputting predictions with minimal delay. Once integrated with lightweight neural networks or incremental learning techniques \cite{bifet2023machine, montiel2021river, lara2023data} at the system level, this method may overcome the high-latency bottleneck of traditional batch-based inference. Moreover, tree models (e.g., Hoeffding trees), online linear models \cite{montiel2021river}, and adaptive ensemble approaches can handle concept drift in dynamic environments. Finally, incorporating physiological signals \cite{saez2021gesture, bimbraw2024forearm, kim2023emg} or multimodal information \cite{rahim2024advanced, yaseen2024next, shen2021simulating} may further enhance overall performance while enabling early recognition.

\section{DUO STREAMERS}

\subsection{Stream-Based RNN-lite Model}

Our framework redesigns certain mechanisms of the RNN architecture to overcome the limitations of mainstream gesture recognition models that rely on high-performance GPUs and cannot be easily deployed on edge devices for streaming recognition. Merely improving computational efficiency does not suffice; the model itself must be small enough to be truly suitable for edge devices.

To this end, we propose a novel RNN-lite architecture. By storing temporal information in a highly compressed external hidden state, this architecture no longer retains large amounts of sequence-related weights—only those needed to process the current input:  

\begin{equation}
\label{eq:external hidden state}
\bigl(h_{t}^{(\mathrm{ext})}\bigr) \;=\; \mathrm{RNNLITE}\Bigl(\bigl(h_{t-1}^{(\mathrm{ext})}\bigr),\, x_t;\, \Theta\Bigr).
\end{equation}

$\mathrm{RNNLITE}$ represents the same state-update equations as a conventional RNN. However, its inputs and outputs are now explicitly defined as the external variables $\bigl(h_{t-1}^{(\mathrm{ext})}\bigr)$. In contrast to other frame-level models that aim solely for low latency, our framework continues to capture temporal information through an external hidden state that specialized components maintain in real time. Moreover, because the external hidden state is readily accessible and the model does not buffer raw inputs, the system can output results immediately without waiting for the gesture to finish or the window to be filled, thereby structurally enabling early recognition. Ultimately, this RNN-lite architecture reduces the parameters required for inference to as low as one-ninth (during high activity) to one-thirty-eighth (during idle periods) of the baseline model, thus supporting unified deployment across multiple platforms.

\begin{figure*}[ht]
  \centering
  \includegraphics[width=\textwidth]{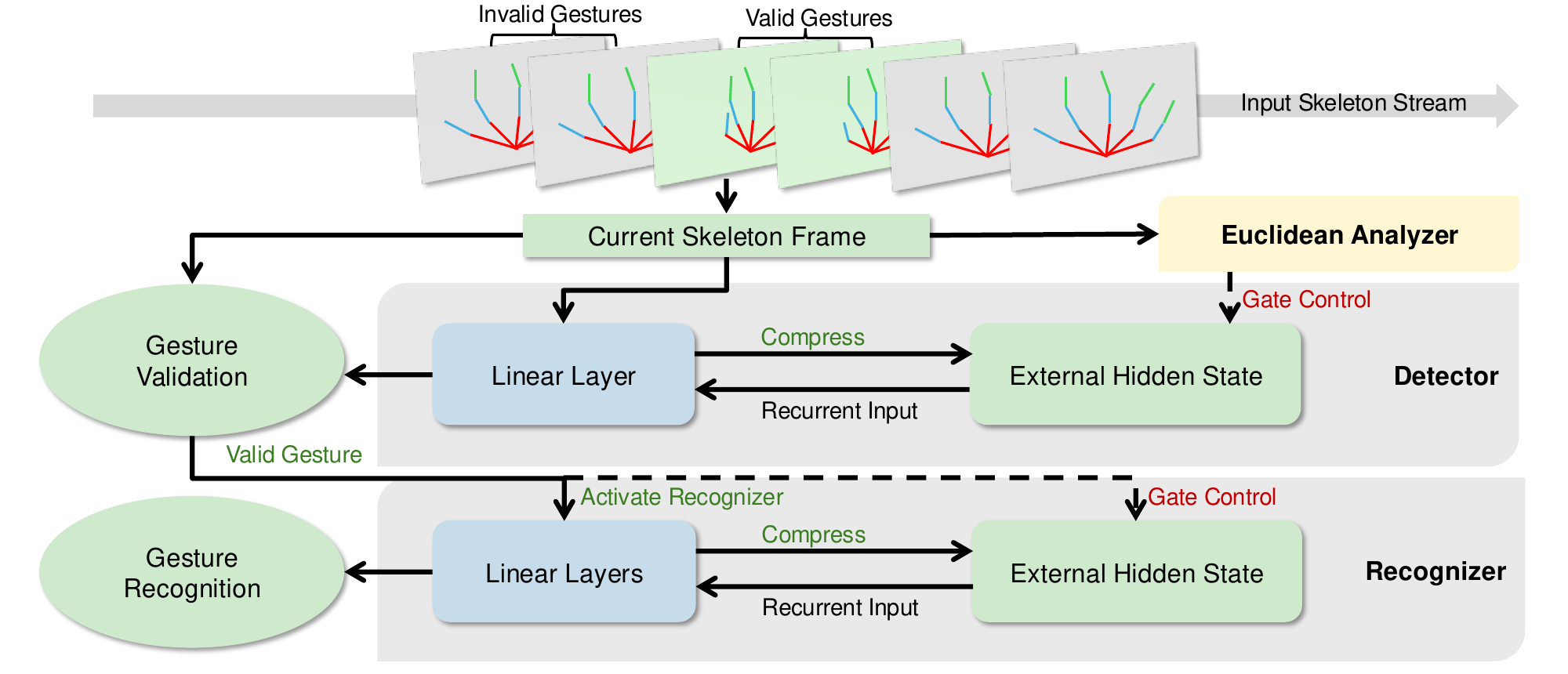}
  \caption{Our proposed RNN-lite streaming model and its three-stage sparse recognition mechanism. The skeleton stream first passes through the Euclidean analyzer and the Detector’s gated external hidden state, continuously monitoring for the presence of valid gestures. Once a valid gesture is detected, the dormant gesture recognition model (Recognizer) is awakened, and during this active phase, gating is applied to the Recognizer’s external hidden state. After the valid gesture concludes, the Recognizer returns to its dormant state, significantly reducing computational overhead and energy consumption. By storing temporal information in a compressed external hidden state and relying solely on the compact Detector for binary classification during idle periods, the model substantially lowers its dependency on hardware resources. Ultimately, the parameters required for inference are reduced to just one-ninth to one-thirty-eighth of the baseline model’s, enabling unified deployment across multiple platforms while enhancing early recognition capabilities.}
  \label{fig:flowchart}
\end{figure*}

\subsection{Three-Stage Sparse Recognition Mechanism}

Gestures commonly used in interactive devices can range from as few as a dozen frames to as many as one or two hundred frames, and they are sparsely distributed among data streams containing thousands of frames. In real-world scenarios, this sparsity can be even more pronounced, meaning that most inputs in a continuous stream are invalid gesture frames—imposing a high resource burden on edge devices that must remain on standby.

To address this sparsity, the Duo Streamers framework does not continuously run a large model. Instead, it uses a Euclidean Analyzer and a small binary classification model (the Detector) to determine whether new inputs appear and whether they are valid. Once valid input is detected, the Detector awakens the dormant gesture recognition model (the Recognizer) and puts it back to sleep when appropriate. By decomposing what would otherwise be handled by a single large model, this approach saves substantial computational resources in streaming scenarios.

Concretely, we design a three-stage sparse recognition mechanism that allows each component to work efficiently, with clear division of labor yet close coordination:
    \begin{enumerate}
        \item Euclidean Analyzer Stage: The raw data stream is first processed by the Euclidean Analyzer, which captures certain joints over serval historical frames and continuously monitors abrupt changes in those joints:  

\begin{equation}
\label{eq:dist-near}
\bar{d}_{\mathrm{near}}(t) \;=\; \frac{1}{N}\,\sum_{k=1}^{N} \bigl\|\, j_{t} \;-\; j_{t-k}\bigr\|,
\end{equation}

\begin{equation}
\label{eq:dist-far}
\bar{d}_{\mathrm{far}}(t) \;=\; \frac{1}{M}\,\sum_{k=N+1}^{\,N+M} \bigl\|\, j_{t} \;-\; j_{t-k}\bigr\|,
\end{equation}

\begin{equation}
\label{eq:reset-condition}
\text{if } \frac{\bar{d}_{\mathrm{near}}(t)}{\bar{d}_{\mathrm{far}}(t)} > \alpha
\quad\text{then}
\quad
h_{t}^{(\mathrm{ext})} \leftarrow \mathbf{0}.
\end{equation}

        By directly gates the external hidden state, this component ensures that the Detector is adequately prepared.
        \item Detector Monitoring Stage: As the data stream flows in continuously, the Detector is always running and listening for valid gestures. When no valid gesture is detected, it carries out minimal computation, conserving overall computational resources and energy. As soon as a valid gesture is identified, the Detector immediately activates the Recognizer.
        \item Recognizer Recognition Stage: Once sparse valid gesture data enters the framework, the Recognizer starts working, while the Detector takes on an additional task—gating the external hidden state updated by the Recognizer to ensure the correct recognition of the current gesture and potential gesture switching. Eventually, once the valid gesture ends, the Detector puts the Recognizer back to sleep and returns to its original monitoring mode, awaiting the next sparse recognition event.
    \end{enumerate}

By employing this three-stage sparse recognition, we eliminate the need for traditional sliding windows to repeatedly invoke large models on consecutive idle frames. Simultaneously, our stream-based RNN-lite model can focus on critical frames, achieving efficient and accurate real-time recognition.

\subsection{Training and Post-Processing Pipeline}
To complement the streaming framework, we developed new training and post-processing pipelines that enable Duo Streamers to be directly trained as a streaming model on commonly used pre-segmented datasets, and further enhance performance at inference through adjustable control mechanisms. Both the Detector and the Recognizer adopt an RNN-lite structure and maintain their own external hidden states. They share the same training procedure: an outer loop sequentially loads each segment, resetting the external hidden state and gradients at the start of every iteration; an inner loop then processes frames one by one and accumulates gradients. After processing an entire segment, the outer loop computes a weighted average gradient and updates the model weights. By performing backpropagation on a per-frame basis, the training pipeline introduces optimizations geared toward early recognition:

\begin{equation}
\label{eq:frame-loss}
\mathcal{L}(\Theta)
\;=\;
\frac{1}{N}
\sum_{n=1}^{N}
\sum_{t=1}^{T_n}
\mathrm{CE}\bigl(\hat{y}_{n,t},\, y_{n,t}\bigr),
\end{equation}

\begin{equation}
\label{eq:weighted-frame-loss}
\mathcal{L}(\Theta)
\;=\;
\frac{1}{N}
\sum_{n=1}^{N}
\sum_{t=1}^{T_n}
w(t)\,\mathrm{CE}\bigl(\hat{y}_{n,t},\, y_{n,t}\bigr),
\end{equation}

On the other hand, in the actual post-processing pipeline, the framework introduces several coefficients for additional performance improvements during deployment:
    \begin{itemize}
        \item A Detector-to-Recognizer activation threshold
        \item A Detector-to-Recognizer deactivation threshold
        \item A minimum waiting time coefficient for the Detector to actively reset the Recognizer’s external hidden state
        \item A Euclidean Analyzer sensitivity coefficient for detecting finger movement
    \end{itemize}

These coefficients affect the recall rate and false alarm rate. They can be flexibly configured to meet different task requirements or applied using robust default values.

In summary, we present three major innovations to fulfill the requirements for streaming gesture recognition. While maintaining comparable performance, we substantially reduce model size and latency, thus enabling streaming data processing and achieving low-latency, real-time gesture recognition across diverse devices.

\section{EXPERIMENTS AND RESULTS}

\begin{figure*}
  \centering
  \begin{subfigure}{0.45\textwidth}
  	\centering
  	\includegraphics[width=\textwidth,height=8cm]{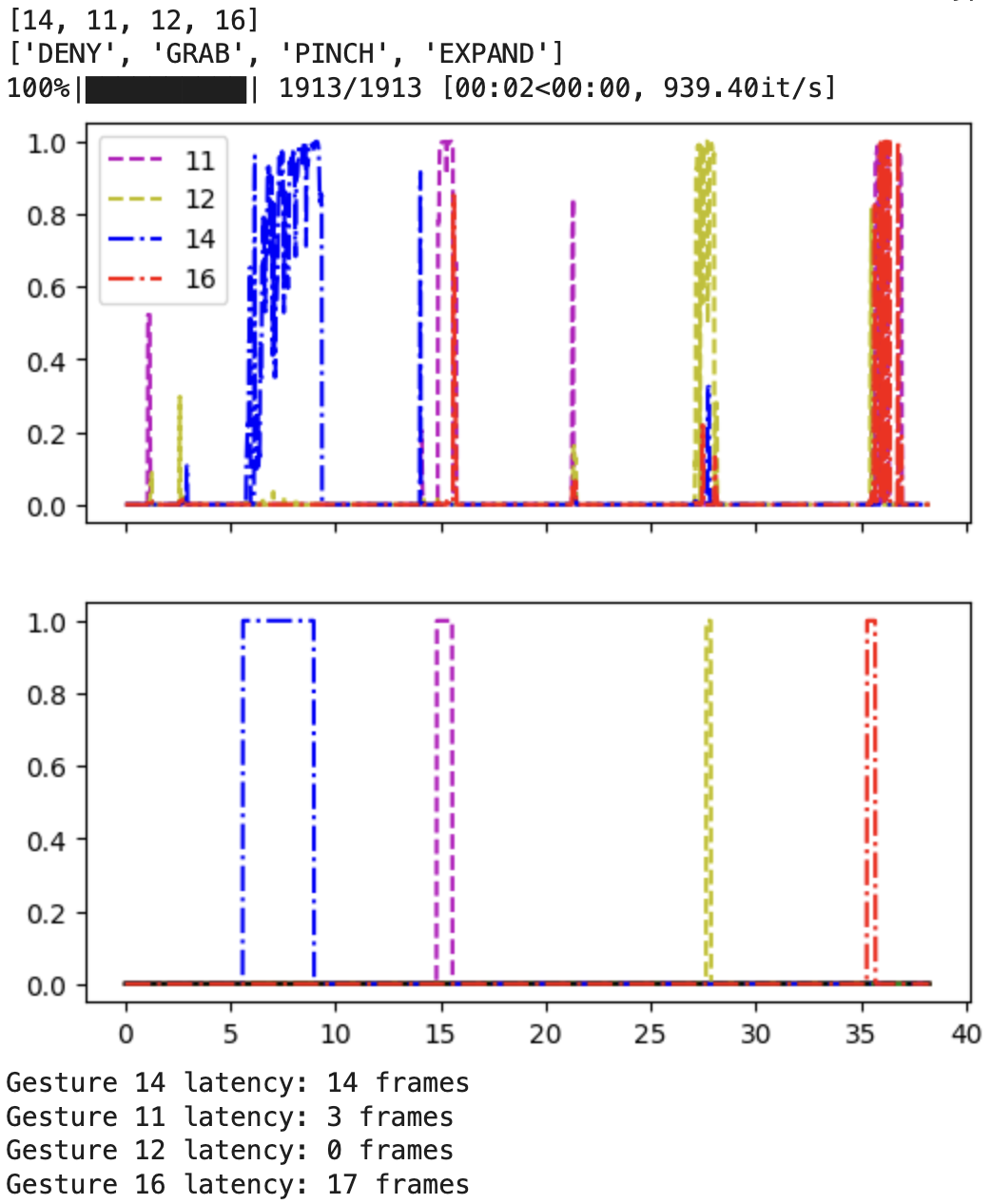}
        \caption{SHREC 2021 skeletal streams}
  	\label{fig:screen_a}
  \end{subfigure}%
  \hfill%
  \begin{subfigure}{0.5\textwidth}
  	\centering
  	\includegraphics[width=\textwidth]{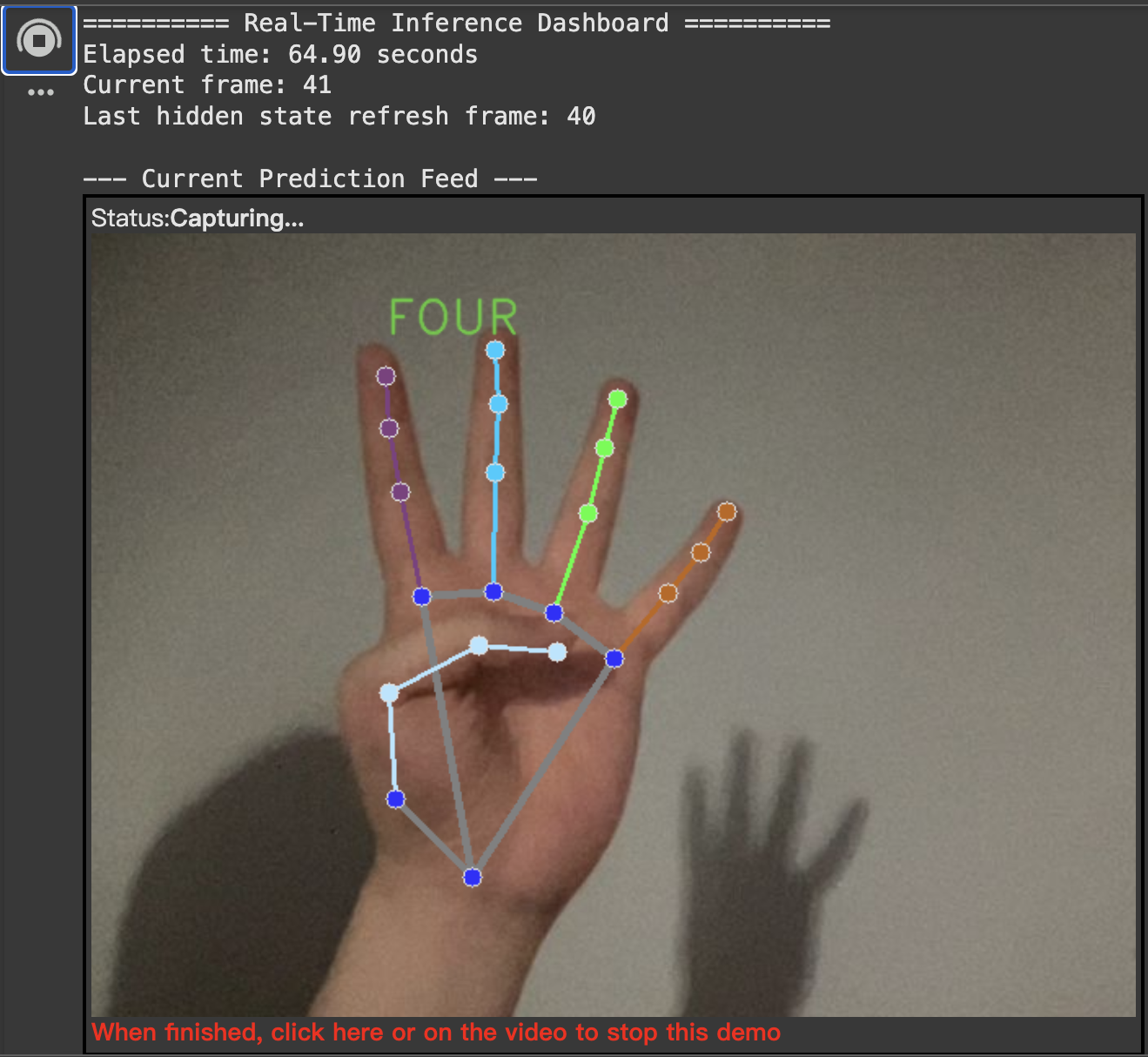} 
        \caption{Camera-based skeletal streams}
  	\label{fig:screen_b}
  \end{subfigure}%
  \\%
  \caption{System logs as it processes data from the SHREC 2021 dataset and camera-based skeletal streams. The left figure presents the detection–recognition outputs and the ground truth, where differently colored dashed lines indicate the confidence of each gesture over time (in seconds), ranging from 0 to 1. The right figure illustrates how the model performs real-time streaming data processing on camera inputs. Given that camera data are pixel-based, we first utilize the Mediapipe library to generate hand keypoint skeleton streams, which are then fed into our framework. The logs indicate that the model balances early recognition with accuracy across various gestures, demonstrating strong adaptability and effectiveness in real-time, high-frequency interactive scenarios. Relevant video content for visualization can be found in the supplementary materials.}
  \label{fig:screenshot}
\end{figure*}

In this section, we first present the implementation details of the Duo Streamers framework, including its core model structures, training procedure, and the hardware used. Next, we describe the dataset employed in our experiments and introduce multiple baseline models for comparative analysis. Finally, we present our results on a series of metrics assessing model accuracy, real-time performance, and size, and we also conduct ablation studies to examine how each component affects overall performance. Our experimental design aims to validate the effectiveness of the three-stage sparse recognition mechanism and the RNN-lite model architecture, as well as to investigate how the framework’s post-processing pipeline further influences performance—ultimately illustrating the superiority of Duo Streamers for real-time gesture recognition on edge devices.

\subsection{Implementation Details}

We developed the Duo Streamers framework in PyTorch and leveraged the TorchTNT library for training and evaluation. During training, CUDA acceleration was enabled, and the external hidden state was explicitly updated and reset within the training loop to avoid excessive overhead; the batch size was set to 256 to balance convergence speed and generalization. For inference, all hardware acceleration was disabled to simulate the sequential processing of streaming data on resource-constrained devices, thus verifying whether the model can sustain real-time performance. During training, we used the Adam optimizer with a learning rate of 0.001 and gradient clipping (restricting the gradient norm to 1.0), and we employed an exponential learning rate scheduler with a gamma value of 0.9 to speed up convergence. On the SHREC2021 dataset, the Recognizer was trained for 300 epochs, and the Detector was trained for 100 epochs. Additionally, the Euclidean Analyzer does not include any neural network components; its parameters were manually tuned on multiple sequences randomly selected from the training set.

\subsubsection{Model Architecture}

The Duo Streamers framework comprises two neural network models—\textbf{Detector} and \textbf{Recognizer}—each equipped with an external hidden state for maintaining temporal information and enabling lightweight inference. On the data stream, a front-end component called the Euclidean Analyzer detects significant hand displacements and continuously gates the Detector’s external hidden state. Once a valid gesture is identified, the Detector wakes the Recognizer and actively gates its external hidden state to enable accurate classification.

\begin{itemize}
    \item Recognizer: Formed by three linear layers combined with an external hidden state, functioning similarly to a three-layer RNN. It processes incoming frames sequentially and stores historical context in the external hidden state, with a dropout rate of 0.2 applied after each linear layer to improve generalization. This design substantially reduces the overall model size while preserving temporal modeling capabilities.
    \item Detector: Employs a single-layer structure along with an external hidden state to decide whether to activate or deactivate the Recognizer, while also gating the Recognizer’s external hidden state. The Detector focuses on identifying valid gestures and minimizing power consumption during idle frames. Notably, our experiments indicate that replacing this single-layer structure with a single-layer LSTM (processing skeletal data recurrently) can significantly enhance the Detector’s ability to delineate valid gesture boundaries.
\end{itemize}

\subsubsection{Hardware}

We conducted all training experiments on Google Colaboratory using an NVIDIA L4 GPU for acceleration. During inference, we switched to a CPU platform to simulate typical edge or embedded device conditions. Owing to the compact design of the Detector and Recognizer, coupled with the three-stage sparse recognition mechanism’s efficient allocation of computational resources, Duo Streamers still achieves a satisfactory level of real-time performance under these conditions.

\subsection{Datasets}

The SHREC2021 dataset contains 180 skeletal gesture sequences recorded in natural environments, with 110 sequences for training and 70 for testing, covering a total of 17 gesture categories \cite{caputo2021shrec}. No preprocessing operations—such as normalization, standardization, or denoising—were applied. One key advantage of the Duo Streamers framework is that it can be trained directly on batch-processed data while still performing streaming inference, thereby simplifying the data preparation process.

\subsection{Baseline Models}

To validate the effectiveness of the Duo Streamers framework for gesture recognition tasks, we selected three models highlighted in the SHREC2021 paper as our primary baselines: a frame-based CNN model \cite{caputo2021shrec}, a window-based model adopting a Transformer architecture \cite{caputo2021shrec}, and a window-based model employing GRU/LSTM modules \cite{caputo2021shrec}.

We chose these models for three main reasons: (1) CNN, RNN, and Transformer architectures encompass common approaches in gesture recognition, and we deliberately avoided methods requiring extensive handcrafted preprocessing, ensuring that our baseline models are both representative and broadly applicable; (2) SHREC2021 provides complete comparative data for these three categories of models, making it convenient for structured experiments and precise measurements; and (3) although some new studies have reported progress on this dataset since 2023, such advances typically rely on additional pre-/post-processing steps or substantially larger network sizes, rather than achieving breakthroughs within these foundational frameworks, thus complicating any rigorous performance comparisons. 

Because SHREC2021 does not release the source code for these baselines, we reproduced three representative models: (1) a window-based lightweight hybrid model that uses a CNN for feature extraction and significantly reduces the parameter count while combining LSTM and residual modules \cite{huang2024video}; (2) a window-based model adapted from the SHREC2021-Transformer approach \cite{caputo2021shrec}; and (3) a window-based, low-latency hybrid model derived from Gesture Spotter, integrating LSTM and self-attention \cite{9873969}. 

By conducting a parallel comparison of these baseline models, we aim to demonstrate two points: first, that Duo Streamers achieves accuracy metrics comparable to mainstream frameworks; and second, that it significantly outperforms traditional solutions in terms of real-time performance and model size, underscoring its potential viability for high-interaction, resource-constrained scenarios.

\subsection{Results on Evaluation Metrics}

\begin{table*}[ht]
  \centering
  \caption{Online Gesture Recognition on SHREC2021. The table presents detection rate, false positive rate, and Jaccard index for various gesture recognition methods under different experimental runs. Frame-Based and Window-Based approaches originate from different base models and their variants \cite{caputo2021shrec}, whereas Stream-based (Ours) is the proposed streaming method. The results show that, compared with conventional methods, the streaming model notably reduces model size and enhances real-time capability while maintaining comparable accuracy.}  
  \begin{tabular}{llccc}
    \toprule
    \textbf{Method} & \textbf{Runs} & \textbf{Detection Rate} & \textbf{False Positives Rate} & \textbf{Jaccard Index} \\
    \midrule
    Frame-Based (CNN) \cite{caputo2021shrec} 
    & mean $\pm$ std 
    & 0.4896 $\pm$ 0.0050 
    & 0.5469 $\pm$ 0.5377 
    & 0.3615 $\pm$ 0.1192 \\
    \midrule
    Window-Based (Transformer) \cite{caputo2021shrec}
    & mean $\pm$ std
    & 0.7060 $\pm$ 0.0212
    & 0.3889 $\pm$ 0.1501
    & 0.5422 $\pm$ 0.0650 \\
    \midrule
    Window-Based (GRU \& TSGR) \cite{caputo2021shrec}
    & mean $\pm$ std
    & 0.7014 $\pm$ 0.0845
    & 0.3044 $\pm$ 0.0348
    & 0.5806 $\pm$ 0.0710 \\
    \midrule
    \multirow{3}{*}{Stream-based \textbf{(Ours)}}
    & Run 1 & 0.7076 & 0.6818 & 0.5356 \\
    & Run 2 & 0.6971 & 0.6735 & 0.5248 \\
    & Run 3 & 0.6812 & 0.6012 & 0.4863 \\
    \bottomrule
  \end{tabular}
  \label{tab:online_comparison}
\end{table*}

\subsubsection{Online Gesture Recognition}

For the online gesture recognition task, we employed five metrics to evaluate model performance. In particular, the Jaccard Index measures the intersection over union (IoU) between the model’s predicted gesture sequences and the ground truth in the time dimension \cite{bras2018gesture}, while the Detection Rate determines whether the model’s prediction is correct and achieves at least a 0.5 IoU with the ground truth \cite{caputo2021shrec, chen2023gesture}. The False Positive Rate is the ratio of the number of incorrectly predicted gestures to the total number of gestures for each category in the sequence \cite{chen2023gesture}. Model Parameters reflect the computational and storage resources required during both training and inference \cite{mantecon2016hand}. Finally, the Real-Time Factor is the ratio of the actual time taken to process the entire test set to the theoretical time, where the latter is derived from the total number of frames divided by the standard frame rate of 50 frames per second \cite{cheng2009hardware, jaramillo2020real}.

Further, we follow the standard procedure below to closely approximate a real deployment environment and conduct a unified evaluation of the models:

\begin{enumerate}
    \item GPU acceleration is disabled.
    \item Parallel processing of sequences is prohibited; the model must process all sequences in order.
    \item After processing each sequence, the model outputs a list containing confidence scores, predicted gesture positions, and predicted gesture categories.
\end{enumerate}

As an accuracy metric outcome, detection rate, Jaccard index, and false positive rate for different runs are summarized in Table~\ref{tab:online_comparison}. As shown in Table~\ref{tab:online_comparison}, the Duo Streamers framework still achieves accuracy levels comparable to the baseline models under resource-constrained conditions.

Detailed experimental results concerning model size and real-time performance metrics are documented in Table ~\ref{tab:model_parameters} and Table~\ref{tab:real_time_performance}. Particularly, the Real-Time Factor is a critical criterion for determining the model’s ability to operate in real-time: a lower Real-Time Factor indicates that the model can process inputs more quickly and can be deployed on more lightweight devices. As indicated by the tables, the Duo Streamers framework outperforms the baseline models in real-time performance while significantly reducing model size.

\begin{table}[H]
    \centering
    \caption{Comparison of the trainable parameter scale of each model in the SHREC2021 online gesture recognition task. The table lists the number of trainable parameters of the three baseline models we reproduced and the Duo Streamers proposed in this study. Among them, "Duo Streamers (Ours) - Idle" indicates the idle state parameter scale when the system is mostly run by only the Detector and the Euclidean Analyzer, and "Duo Streamers (Ours) - Busy" indicates the maximum parameter scale when the Detector and the Recognizer are working at high frequency at the same time.}
    \label{tab:model_parameters}
    \begin{tabular}{lc}
        \toprule
        \textbf{Model} & \textbf{Trainable Parameters} \\
        \midrule
        CNN+LSTM+ResNet \cite{huang2024video}  & 6{,}136{,}771 \\
        SHREC2021 - Transformer \cite{caputo2021shrec}      & 12{,}698{,}129 \\
        Gesture Spotter - LSTM+Attention \cite{9873969}    & 27{,}759{,}633 \\
        \textbf{Duo Streamers (Ours) - Idle}              & \textbf{408{,}066}     \\
        \textbf{Duo Streamers (Ours) - Busy}           & \textbf{1{,}800{,}723}  \\
        \bottomrule
    \end{tabular}
\end{table}

\begin{table}[H]
    \centering
    \caption{Comparison of the real-time processing capabilities of various models in online gesture recognition. The Real-Time Factor (RTF) measures the ratio between a model’s inference time and the theoretical input sequence duration. An RTF below 1 indicates that the model can complete real-time inference at the standard frame rate. Importantly, as gesture recognition algorithms are deployed on a wide range of edge devices, a smaller RTF enables real-time inference on more compact hardware. The table presents three reproduced baseline models alongside our proposed Duo Streamers, demonstrating that Duo Streamers delivers superior real-time performance.}
    \begin{tabular}{lc}
        \toprule
        \textbf{Model} & \textbf{Real-Time Factor} \\
        \midrule
        CNN+LSTM+ResNet \cite{huang2024video} & 0.8335 \\
        SHREC2021 - Transformer \cite{caputo2021shrec}      & 1.0810 \\
        Gesture Spotter - LSTM+Attention \cite{9873969}   & 0.5165 \\
        \textbf{Duo Streamers (Ours)}                & \textbf{0.0627} \\
        \bottomrule
    \end{tabular}
    \label{tab:real_time_performance}
\end{table}

Additionally, the Duo Streamers framework achieved an Early Detection Latency of 6.38 frames on the SHREC2021 dataset. This latency is shorter than the shortest gesture length in the dataset, indicating that for most gestures, the framework can return predictions in the early stages of execution.

We experimented with different threshold settings in the framework’s post-processing pipeline to examine their effects on accuracy metrics. We found that a lower activation threshold tended to increase the detection rate but could potentially raise the false positive rate, whereas a higher activation threshold helped reduce the false positive rate but might lead to a decrease in detection rate.

For instance, with the Recognizer activation threshold initially set to 0.45 and the deactivation threshold at 0.2, the online recognition results yielded a detection rate of 0.7077 and a false positive rate of 0.6818. When the thresholds were adjusted to a more stringent level—setting the activation threshold to 0.8 and the deactivation threshold to 0.5—the detection rate decreased to 0.6812 while the false positive rate improved to 0.6012. Therefore, we recommend raising the activation threshold as much as possible, without incurring excessive loss in detection rate, to better control false positives.

\subsubsection{Ablation Studies}

To assess the contribution of individual components within the Duo Streamers framework, we conducted ablation experiments focusing on three key elements:

\begin{enumerate}
    \item \textbf{External Hidden State}: We evaluated the model’s ability to capture temporal dependencies when the external hidden state was removed.
    \item \textbf{Gating Mechanism}: We investigated the impact on overall performance in the absence of the control mechanism for the external hidden state.
    \item \textbf{Detector Model}: We examined how modifications or removal of the Detector model affected the false positive rate.
\end{enumerate}

The results of these experiments are summarized in Table~\ref{tab:ablation_studies_part1}.

\begin{table}
\centering
\caption{Ablation study of the key components in Duo Streamers. Rows indicate the specific component removed from the full framework. The results demonstrate the functions and necessity of each module: the external hidden state is responsible for capturing temporal information; The gating mechanism maintains the external hidden state of the two sub-models; The detector model continuously monitors the stream for gesture inputs; removing this module leads to a significant increase in the recognizer's false positive rate.}
\begin{tabular}{lcc}
\toprule
\textbf{Component Removed} & \textbf{Detection Rate} & \textbf{False Positive Rate} \\
\midrule
\textbf{None (Full Framework)}          & \textbf{0.7077} & \textbf{0.6818} \\
External Hidden State      & 0.5694 & 0.6841 \\
Gating Mechanism          & 0.0835 & --- \\
Detector Model             & --- & 0.9888 \\
\bottomrule
\end{tabular}
\label{tab:ablation_studies_part1}
\end{table}

\subsubsection{Observations and Discussion}

The Duo Streamers framework demonstrates robust performance in online gesture recognition tasks, highlighting its suitability for practical applications involving streaming data. Delving deeper, ablation experiment results confirm that each component of the framework makes a meaningful contribution to overall performance. Overall, despite its ultra-lightweight design, the Duo Streamers framework’s online recognition performance is comparable to that of larger, more complex models, making it particularly suitable for deployment on edge devices or in environments with limited computational resources. 


\subsection{Limitation and Future Work}

Despite Duo Streamers’ demonstrated ability to perform real-time gesture recognition on resource-constrained devices, some limitations remain. In complex scenarios, its capacity to precisely identify gesture boundaries requires further improvement, and additional field testing is needed to refine the switching strategy among the three stages while balancing sensitivity and power consumption on edge devices. In the future, we plan to explore more real-time applications and multimodal inputs—such as images, depth data, and skeletal keypoints—and introduce adaptive post-processing and meta-learning methods so that the system can automatically fine-tune sensitivity coefficients and other parameters after deployment. This approach will help extend the ultra-lightweight design to more challenging interaction modes.

\section{CONCLUSION}

Duo Streamers integrates a three-stage sparse recognition mechanism, an RNN-lite streaming model, and a custom training and post-processing pipeline to achieve real-time, lightweight gesture recognition across multiple platforms. In particular, the three-stage sparse recognition effectively reduces redundant computation during idle frames, while the external hidden state preserves contextual information with minimal overhead. Meanwhile, the training and post-processing pipelines designed for streaming applications ensure stable model performance in real-world settings. Experimental results indicate that Duo Streamers achieves accuracy comparable to mainstream methods on datasets collected from real environments, while significantly reducing model size and accelerating inference. Taken together, Duo Streamers provides an efficient solution that balances accuracy and portability, laying a solid foundation for the adoption of gesture recognition in edge computing and wearable device scenarios.




\section*{ETHICAL IMPACT STATEMENT}
This research utilized the publicly available SHREC2021 gesture skeleton dataset, which does not contain facial images, audio, or other personally identifiable information. According to SHREC2021’s public documentation, the dataset providers have anonymized or de-identified the data to minimize potential privacy risks. Moreover, they have reportedly obtained informed consent from all participants during data collection and curation. Throughout our research, we strictly adhered to the dataset license and relevant privacy regulations, refraining from any commercial or surveillance applications beyond the research scope and avoiding any further collection of personal information or facial imagery.

{\small
\bibliographystyle{ieee}
\bibliography{egbib}
}

\end{document}